\pdfoutput=1

\documentclass[11pt]{article}

\usepackage[]{naacl2021}

\usepackage{times}
\usepackage{latexsym}

\usepackage{multirow}
\usepackage[multiple]{footmisc}
\usepackage{array, booktabs, makecell}
\usepackage{graphicx}
\usepackage{colortbl}
\usepackage{xcolor}
\usepackage{xurl}
\usepackage{breakurl}
\usepackage{microtype}

\usepackage{amsmath}
\usepackage{amssymb}
\usepackage{amsfonts}
\usepackage{multirow}
\usepackage{arydshln}
\usepackage{enumitem}

\newcommand{\B}[1]{\mathbf{#1}}

\newcommand{\tb}[1]{\textbf{#1}}
\newcommand{\ti}[1]{\textit{#1}}
\newcommand{\Cal}[1]{\mathcal{#1}}
\newcommand{\Sref}[1]{\S\ref{#1}}

\newcommand{\cmu}{$^1$}
\newcommand{\salesforce}{$^2$}

\definecolor{darkgreen}{rgb}{0.0, 0.42, 0.24}
\usepackage[T1]{fontenc}

\usepackage[utf8]{inputenc}

\usepackage{microtype}

\title{Focused Attention Improves Document-Grounded Generation}

\author{Shrimai Prabhumoye\cmu{}\thanks{~~~Work done during internship at Salesforce.} , Kazuma Hashimoto\salesforce{}, Yingbo Zhou\salesforce{}, \\ \bf Alan W Black\cmu{}, Ruslan Salakhutdinov\cmu{} \\
\cmu{}Carnegie Mellon University, \salesforce{}Salesforce Research\\
  \texttt{\{sprabhum@cs.cmu.edu, k.hashimoto@salesforce.com\}} \\
}

\begin{document}
\maketitle
\begin{abstract}
Document grounded generation is the task of using the information provided in a document to improve text generation.
This work focuses on two different document grounded generation tasks: \emph{Wikipedia Update Generation} task and \emph{Dialogue response generation}.
Our work introduces two novel adaptations of large scale pre-trained encoder-decoder 
models focusing on building context driven representation of the document and enabling specific attention to the information in the document.
Additionally, we provide a stronger BART baseline for these tasks.
Our proposed techniques outperform existing methods on both automated (at least $48$\% increase in BLEU-4 points) 
and human evaluation for closeness to reference and relevance to the document.
Furthermore, we perform comprehensive manual inspection of the generated output and categorize errors to provide insights into future directions in modeling these tasks.
\end{abstract}

\section{Introduction}
\label{sec:gg_intro}

Natural language generation (NLG) systems are increasingly expected to be naturalistic, contentful, and situation-aware due to their popularity and pervasiveness in human life~\cite{reiter2000buildNLG,ws-2014-international-natural}.
This is particularly relevant in dialogue systems~\cite{zhang2018personalizing,niu-bansal-2018-polite}, machine translation systems~\cite{mirkin-meunier-2015-personalized,rabinovich-etal-2017-personalized}, story generation~\cite{fan-etal-2018-hierarchical,yao2019plan}, and question answering systems~\cite{gatius-2017-personalized,reddy-etal-2019-coqa}.

Despite these mainstream applications,  NLG systems face the challenges of being bland, devoid of content, generating generic outputs and hallucinating information~\cite{wiseman:2017,li:2015,Holtzman2020The,Welleck2020Neural}.
Grounding the generation in different modalities like images~\cite{huang2016visual,mostafazadeh-etal-2017-image,image_chat}, videos~\cite{palaskar-etal-2019-multimodal,regneriGroundedAction}, and structured data~\cite{banik-etal-2013-kbgen,gardent-etal-2017-webnlg} alleviates some of these issues.
Generating natural language from schematized or structured data such as database records, slot-value pair, and Wikipedia Infobox has been explored in prior work~\cite{mei2016talk,sem_cond_lstm,wikibio}. 
Although useful, these tasks encounter difficulties such as general applicability (databases may not be available for all domains) and are constrained by the available resources (size of the database).

Document grounded generation mitigates these applicability issues by exploiting the vast availability of data in unstructured form (e.g. books, encyclopedias, news articles, and Wikipedia articles).
This enhances the applicability of document grounded generation to a wide range of domains with limited (or no) availability of structured data.
Hence, recent work has focused on defining new tasks and carving the scope of the problems~\cite{WikiGen2018,prabhumoye-etal-2019-towards,faltings2020text,zhou-etal-2018-dataset,dinan2018wizard}.


\begin{figure*}[t]
\centering
\includegraphics[scale=0.35]{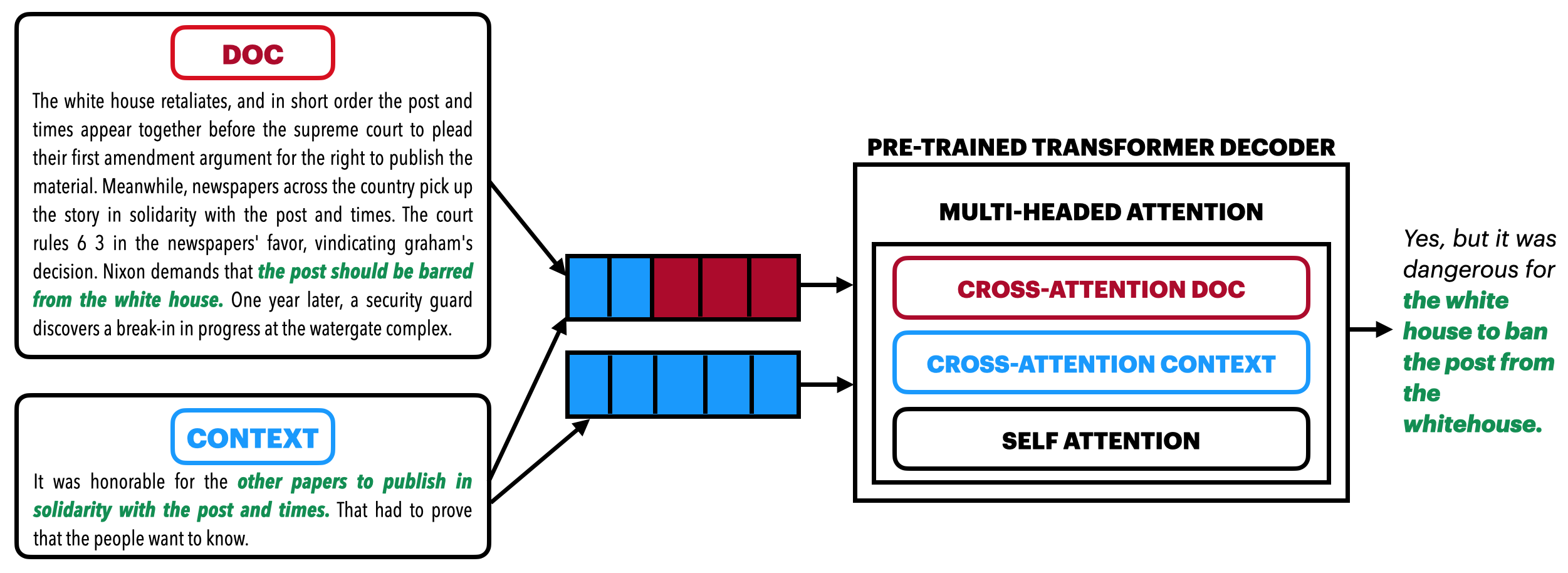}
\vspace{-0.5em}
\caption{Document Grounded Generation: An example of a conversation that is grounded in the given document (text in \textcolor{darkgreen}{green} shows information from the document that was used to generate the response).}
\label{fig:gg_overview}
\vspace{-1em}
\end{figure*}

We focus on two different document grounded generation tasks: (1) Wikipedia Update Generation task~\cite{prabhumoye-etal-2019-towards} and (2) Dialogue response generation~\cite{zhou-etal-2018-dataset,dinan2018wizard}.
Prior work has studied these two tasks independently and focused on task specific modeling techniques~\cite{Zhao2020Low-Resource,zhao2020knowledgegrounded,prabhumoye-etal-2019-towards}.
Our work unifies these tasks and formally shows the similarity in them: presence of a context and a document to ground the information in the generation process.

Our work introduces two novel improvements to the architectures of large scale pre-trained models~\cite{lewis2019bart,raffel2019exploring}: (1) we focus on building context driven representation of the document, where the context is taken into account while building the representation of the document, and (2) during generation we provide specific attention to the information in the document.
We provide a stronger BART-based~\cite{lewis2019bart} baseline for these tasks. 
This work shows that pre-trained models albeit good at text generation, can be further improved by providing grounding specific improvements.

Our main contributions are the two new proposed techniques for the document grounded generation tasks (\Sref{sec:gg_codr} and \Sref{sec:gg_doha}).
We also provide a new baseline which is stronger than the previous state-of-the-art methods~\cite{zhao2020knowledgegrounded, prabhumoye-etal-2019-towards} for the two tasks.
We formally show how the two independent tasks studied in this paper are identical and similar modeling techniques can be used to solve them (\Sref{sec:gg_method}).
Automated and human evaluation results on three different datasets demonstrate substantial improvements (\Sref{sec:gg_auto_eval} and \Sref{sec:gg_human_eval}).
Specifically, we achieve an improvement of $19.7$ BLEU-4 points compared to~\citet{zhao2020knowledgegrounded} on the dialogue generation task.
Additionally, significant gains are observed in BLEU-4 compared to BART-based baseline.
A comprehensive manual analysis of the generated output is presented in this work which paves way for future work (\Sref{sec:gg_discuss}).
We will release our code on \href{https://github.com/shrimai/Focused-Attention-Improves-Document-Grounded-Generation}{Github}. 



\section{Task Definition}
\label{sec:gg_formal_task}

Our task is to generate text given a context and a source of content (document).
Additionally, the generated text should coherently fit the context and contain information from the document.
We focus on content present in unstructured form in documents to ground text generation.
Figure~\ref{fig:gg_overview} illustrates such an example.
Dialogue response generation is traditionally conditioned on the dialogue context~\cite{vinyals2015,li:2015}.
As Figure~\ref{fig:gg_overview} demonstrates, the generative model is conditioned on both the document as well as the dialogue context.
Note that the context and document play different roles in impacting the generation -- the context sets the background while the document provides the content necessary to generate the text.

Formally, each sample $\B{i}$ of our task is defined as a tuple ($\B{d}_i, \B{c}_i, \B{x}_i$) containing context $\B{c}_i$, document $\B{d}_i$ and text $\B{x}_i$ to be generated. 
Note that each $\B{d}_i$ can be a single document or a set of documents.
The task is to generate $\B{x}_i$ such that it coherently follows $\B{c}_i$ and contains information from $\B{d}_i$. 
The task can be modeled as the following conditional text generation model: 
$p_\theta(\B{x}_i | \B{c}_i, \B{d}_i),$
where $\theta$ is a set of model parameters.

Figure~\ref{fig:gg_overview} illustrates that the generator has to account for two inputs the dialogue context $\B{c}_i$ (shown in blue) and the document $\B{d}_i$ (shown in red) to generate the response $\B{x}_i$ grounded in $\B{d}_i$ (text shown in green).
If the generative model was only conditioned on dialogue context, then it could produce generic responses like ``\ti{Do you think they did the right thing?}'' or ``\ti{Yes, I agree.}'' or hallucinate information like ``\ti{Yes, and the Times published it on the front page.}''.
These which would be appropriate to the given context but are devoid of content or contain wrong information. 
Document grounded models are capable of responding with interesting facts like ``\ti{Yes, but it was dangerous for the white house to ban the post from the white house.}''

\section{Methodology}
\label{sec:gg_method}

A natural way to model $p_\theta(\B{x}_i | \B{c}_i, \B{d}_i)$ is to train an encoder-decoder model using cross-entropy loss $-\log{p_\theta}$ with respect to the ground-truth output text.
We discuss two ways of building effective representations for encoder-decoder models to focus on $\B{d}_i$: (1) combine encoder representations of $\B{c}_i$ and $\B{d}_i$, (2) include an additional attention multi-head at each layer of the transformer to specifically focus on the content in $\B{d}_i$.

\subsection{Baselines}

\paragraph{Low-Res:}
\citet{Zhao2020Low-Resource} introduce the state-of-the-art model for document grounded dialogue generation.
As described in (\Sref{sec:gg_formal_task}), the chat history serves as the context $\B{c}_i$ and $\B{x}_i$ is the response to be generated.
\citet{Zhao2020Low-Resource} pre-train their architecture on the dialogue specific Reddit~\cite{dziri2018augmenting} dataset and learn separate parameters for encoding $\B{c}_i$ and $\B{d}_i$.
\citet{Zhao2020Low-Resource} further has three components--context processor, knowledge processor and the language model, each of which build distributions over the vocabulary space.
A decoding manager is then trained to generate a token based on these three distributions.

Instead, we employ the recent success of the pre-trained encoder-decoder models~\cite{lewis2019bart,raffel2019exploring} by using BART~\citep{lewis2019bart}.
One key component of solving this task is to build a representation of the content in the document/s $\B{d}_i$ that is \emph{not} present in the context $\B{c}_i$.
We want to leverage the \emph{SelfAttention} feature of transformers~\cite{vaswani2017attention} to build such a representation.
Since, we use a pre-trained language model as our baseline architecture, we don't use a separate language model component.
Instead, we direct our efforts to focus on effectively combining $\B{c}_i$ and $\B{d}_i$.

\paragraph{Content Transfer:}
\citet{prabhumoye-etal-2019-towards} provide benchmark numbers for the Wikipedia Update Generation task (\Sref{sec:gg_formal_task}).
They explore multiple generative as well as extractive models with and without context.
We use their best performing Context Informed LSTM-based encoder-decoder model as baseline.
This model concatenates the tokens of the context $\B{c}_i$ and the document $\B{d}_i$ and passes the concatenated sequence to the encoder.


\paragraph{BART:} The most straightforward way of using BART for modeling $p_\theta(\B{x}_i | \B{c}_i, \B{d}_i)$ is to concatenate the tokens of the context $\B{c}_i$ and the document $\B{d}_i$ and pass the concatenated sequence ($[\B{c}_i; \B{d}_i]$) to the BART encoder, and then the decoder generates $\B{x}_i$.
This is our BART baseline; it already has the advantage of the highly contextualized representations of $\B{c}_i$ and $\B{d}_i$ in comparison with \citet{Zhao2020Low-Resource}.
However, fully relying on the self-attention mechanism over the concatenated text would lack the explicit distinction between $\B{c}_i$ and $\B{d}_i$.

Below, we describe two techniques to efficiently build document focused representations.
In Figure~\ref{fig:gg_overview}, the method which adds an additional \emph{CrossAttention} multi-head sub-layer to each layer of the transformer is shown. 
This attention multi-head specifically focuses on the document $\B{d}_i$. 

\subsection{Context Driven Representation}
\label{sec:gg_codr}
%

We propose to use two encoder representations for $\B{c}_i$ and $\B{d}_i$.
We first define $\B{h}_d = \mathsf{Encoder}([\B{c}_i; \B{d}_i])$ to get a contextualized representation of $\B{d}_i$, conditioning on the context $\B{c}_i$.
$\B{h}_d$ is equivalent to the representation used in the BART baseline.
We then apply the same BART encoder to the context alone: $\B{h}_c = \mathsf{Encoder}(\B{c}_i)$.
We finally concatenate the encoder outputs $\B{h} = [\B{h}_c; \B{h}_d]$ before passing them to the BART decoder.
This $\B{h}$ is \tb{Co}ntext \tb{D}riven \tb{R}epresentation (\tb{CoDR}).
This method does not require any model architectural modification, and instead the encoder and decoder are fined-tuned to use the multiple input representations.

\subsection{Document Headed Attention}
\label{sec:gg_doha}

In this section, we describe \tb{Do}cument \tb{H}eaded \tb{Attention} (\tb{DoHA}) to further enhance the use of the multiple input representations.
A decoder in transformer encoder-decoder models~\citep{vaswani2017attention} has two types of multi-head attention mechanism, SelfAttention and CrossAttention with the source sequence.
\emph{SelfAttention} module allows each position in the decoder to attend to all positions in the decoder up to and including that position.
\emph{CrossAttention} module performs multi-head attention over the output of the encoder stack and attends over the source sequence.
While our CoDR method uses the two different source representations, $\B{h}_c$ and $\B{h}_d$, \emph{CrosstAttention} is still shared over the concatenated representation $\B{h}$.

In this work, we add an additional multi-head attention \emph{CrossAttention\_Doc} to specifically attend over the tokens of the document, while the original \emph{CrossAttention} (named as \emph{CrosstAttention\_Cxt}), only attends over the tokens of the context.
Each of the multi-heads are of the form:
\begin{eqnarray*}
\mathsf{MultiHead}(Q, K, V) = [\B{H}_1; \ldots; \B{H}_m]\B{W^{o}}, \\
\B{H}_j = \mathsf{Attention}(Q\B{W^{Q}_{j}}, K\B{W^{K}_{j}}, V\B{W^{V}_{j}}).
\end{eqnarray*}
The multi-head function receives three inputs - a query $Q$, key $K$ and value $V$.
$\B{W^{o}}$ is an output projection of the concatenated outputs of the attention heads.
Each $\B{H}_j$ is the output of a single attention head and $\B{W^{Q}_{j}}$, $\B{W^{K}_{j}}$ and $\B{W^{V}_{j}}$ are head-specific projections for $Q$, $K$, and $V$, respectively.

Hence, the multi-head \emph{CrossAttention\_Doc} is defined by:
\begin{eqnarray*}
\mathsf{CrossAttention\_Doc}(Q, K, V) \\ = [\B{H}_1; \ldots; \B{H}_m]\B{W^{do}}, \\
\B{H}_j = \mathsf{Attention}(Q\B{W^{dQ}_{j}}, K\B{W^{dK}_{j}}, V\B{W^{dV}_{j}}),
\end{eqnarray*}
where $\B{W^{do}}, \B{W^{dQ}_{j}}, \B{W^{dK}_{j}}$ and $\B{W^{dV}_{j}}$ are parameters trained specifically to focus on document.
The parameters of \emph{CrossAttention\_Doc} are initialized with those of \emph{CrossAttention\_Cxt}.

Each decoder layer follows the following sequence of functions:
\begin{eqnarray*}
\B{h}&=&\Cal{F}(\mathsf{SelfAttention}(\B{h_x}, \B{h_x}, \B{h_x})), \\
\B{h}&=&\Cal{F}(\mathsf{CrossAttention\_Cxt}(\B{h}, \B{h_c}, \B{h_c})), \\
\B{h}&=&\Cal{F}(\mathsf{CrossAttention\_Doc}(\B{h}, \B{h_d}, \B{h_d})), \\
\B{h}&=&\Cal{F}(\mathsf{FFN}(\B{h})),
\end{eqnarray*}
where $\Cal{F}(\B{h})$ is a sequence of $\mathsf{LayerNorm}(\mathsf{residual} + \mathsf{dropout}(\B{h}))$, followed by $\mathsf{residual} = \B{h}$.
We integrate the additional attention head \emph{CrossAttention\_Doc} by passing the output of the previous attention head \emph{CrossAttention\_Cxt} as query.
Unlike the weighted attention fusion techniques~\cite{cao-etal-2020-pretrained}, this technique of fusing the additional attention head is novel and useful as it does not require any additional parameters for the fusion.

\section{Document Grounded Generation Tasks}
\label{sec:gg_tasks}

Document grounded generation can leverage unstructured data as a source of grounding and can hence be applied to a variety of generation tasks such as dialogue responses, Wikipedia articles, reports and legal argument.
This work focuses on \emph{Wikipedia Update Generation} and \emph{Dialogue Response Generation} which have been studied independently in prior work.
We discuss the similarities in these two tasks and design a common modeling technique for them.

\begin{table*}[t]
\centering
\small{
\begin{tabular}{@{}l r r r r r r r@{}}
\textbf{Model} & \textbf{BLEU-1} & \textbf{BLEU-2} & \textbf{BLEU-3} & \textbf{BLEU-4} & \textbf{Rouge-L} & \textbf{Meteor} & \textbf{F1}\\
\toprule
\multicolumn{8}{c}{\textbf{Wikipedia Update Generation}} \\
\midrule
Content Transfer~\cite{prabhumoye-etal-2019-towards} & 10.18 & 4.42 & 2.20 & 1.23 & 10.08 & 6.21 & 12.6 \\
BART (baseline) & 21.72 & 14.71 & 11.28 & 9.20 & 22.39 & 12.90 & 27.5 \\
CoDR & \tb{25.15} & \tb{17.33} & \tb{13.56} & \tb{11.31} & 23.48 & \tb{14.38} & 29.0 \\
DoHA & 25.11 & 17.04 & 13.17 & 10.86 & \tb{23.49} & 14.28 & 29.1 \\
\toprule
\multicolumn{7}{c}{\textbf{CMU\_DoG}} \\
\midrule
Low-Res~\cite{Zhao2020Low-Resource} & 15.00 & 5.70 & 2.50 & 1.20 & - & - & 10.7 \\
BART (baseline) & 23.78 & 19.27 & 17.66 & 16.91 & 19.30 & 12.59 & 21.7 \\
CoDR & 26.86 & 22.75 & 21.30 & 20.68 & 20.41 & 14.47 & 22.7 \\
DoHA & \tb{27.33} & \tb{23.05} & \tb{21.55} & \tb{20.90} & \tb{20.44} & \tb{14.55} & 22.8 \\
\toprule
\multicolumn{8}{c}{\textbf{Wizard of Wikipedia (Seen)}} \\
\midrule
Low-Res~\cite{Zhao2020Low-Resource} & 21.80 & 11.50 & 7.50 & 5.50 & - & - & 18.0 \\
BART (baseline) & 23.92 & 14.62 & 10.24 & 7.75 & 21.41 & 15.45 & 31.1 \\
CoDR & 24.00 & 14.98 & 10.64 & \tb{8.18} & \tb{21.82} & 15.71 & 31.8 \\
DoHA & \tb{24.14} & \tb{15.08} & \tb{10.68} & \tb{8.18} & 21.76 & \tb{15.89} & 31.8 \\
\toprule
\multicolumn{8}{c}{\textbf{Wizard of Wikipedia (Unseen)}} \\
\midrule
Low-Res~\cite{Zhao2020Low-Resource} & 20.70 & 10.10 & 6.20 & 4.30 & - & - & 16.5 \\
BART (baseline) & 21.88 & 12.54 & 8.44 & 6.23 & 19.14 & 14.03 & 28.2 \\
CoDR & 21.84 & 12.74 & 8.60 & 6.35 & 19.50 & 14.22 & 29.0 \\
DoHA & \tb{22.31} & \tb{13.04} & \tb{8.89} & \tb{6.60} & \tb{19.62} & \tb{14.47} & 29.0 \\
\bottomrule
\end{tabular}
}
\vspace{-0.5em}
\caption{Results on the automated metrics for the three datasets}
\vspace{-1.0em}
\label{tab:gg_main_res}
\end{table*}

\subsection{Wikipedia Update Generation}
\label{sec:gg_wiki_upd_task}
This task involves generating an update for Wikipedia context given a news article~\cite{prabhumoye-etal-2019-towards}.
The dataset was collected by parsing Wikipedia articles and Common Crawl for news articles.
It consists tuples of the form ($\B{d}_i, \B{c}_i, \B{x}_i$), where the grounding document $\B{d}_i$ is the news article which contains information for the reference update $\B{x}_i$. 
$\B{x}_i$ is written by a Wikipedia editor as an update to the Wikipedia context $\B{c}_i$.
The goal of the task is to generate $\B{x}_i$ given the context $\B{c}_i$ and the document $\B{d}_i$.

\subsection{Dialogue Response Generation} 
\label{sec:gg_dialogue_resp_gen}
Goal oriented dialogues have been traditionally grounded in structured sources like slot-value pairs and databases~\cite{wei-etal-2018-airdialogue,rastogi2020towards}.
Open domain dialogue generation on the other hand faces the issue of ``hallucinating'' information~\cite{ghazvininejad2018knowledge}.
Hence we study open domain dialogue generation which is grounded in documents as a source of information.

\paragraph{CMU\_DoG:} The CMU Document Grounded Conversations dataset consists of human-human conversations collected over Amazon Mechanical Turk~\cite{zhou-etal-2018-dataset}.
The conversations are grounded in a document provided to the crowd-workers and focuses only on movies. 
The dataset uses Wikipedia descriptions of movies for grounding the conversations.
The dataset consists tuples of the form ($\B{d}_i, \B{c}_i, \B{x}_i$), where $\B{d}_i$ is a section (or passage) extracted from Wikipedia, $\B{c}_i$ is dialogue history (or context) and $\B{x}_i$ is the reference response.
The response $\B{x}_i$ is grounded in $\B{d}_i$ and coherently follows the conversation $\B{c}_i$.

\paragraph{Wizard of Wikipedia:} This dataset also consists of human-human conversations collected over Amazon Mechanical Turk and are grounded in passages extracted from Wikipedia~\cite{dinan2018wizard}.
These conversations are grounded in a diverse range of topics (totally 1365) which are further split into seen and unseen topics during training and validation.
At each step of the dialogue the wizard has access to a set of passages of knowledge which may be relevant to the given dialogue context.
The dataset is created by retrieving the top $7$ articles (first paragraph only) that are most relevant to the last two turns of dialogue (by wizard and apprentice).
Hence, the dataset consists tuples of the form ($\B{d}_i, \B{c}_i, \B{x}_i$), where $\B{d}_i$ is a list of $7$ passages relevant to the conversation, $\B{c}_i$ is dialogue history (or context) and $\B{x}_i$ is the reference response.

The above three tasks consists tuples of the form ($\B{d}_i, \B{c}_i, \B{x}_i$), where $\B{x}_i$ coherently follows $\B{c}_i$ and is grounded in $\B{d}_i$.
Hence, we can use common modeling techniques (\Sref{sec:gg_method}) for these tasks.
\footnote{Data statistics are shown in Appendix (\Sref{sec:appendixA})}

\section{Experiments and Results}
\label{sec:gg_expts_res}


We implement all our models with the transformers tool~\cite{Wolf2019HuggingFacesTS}, and the details are in \Sref{sec:appendixA}.

\subsection{Automated Evaluation}
\label{sec:gg_auto_eval}

Following prior work~\cite{prabhumoye-etal-2019-towards,Zhao2020Low-Resource}, we evaluate our system-generated sentences against the reference sentences on Rouge-L~\cite{lin2004rouge}, BLEU~\cite{papineni2002bleu} and METEOR~\cite{denkowski2011meteor} metrics.\footnote{We use NLG evaluation toolkit~\cite{sharma2017nlgeval} from \url{https://github.com/Maluuba/nlg-eval}}
Rouge-L measures the longest common subsequence between the generated sentence and the reference, capturing both lexical selection and word order.
METEOR also uses synonyms and stemmed forms of the words in candidate and reference sentences, and thus may be better at quantifying semantic similarities.
Additionally, we present F1 which indicates the unigram overlap between the generated output and the reference sentence.\footnote{We use the code published at \url{https://github.com/facebookresearch/ParlAI/blob/master/parlai/core/metrics.py} to calculate unigram F1.}

Table~\ref{tab:gg_main_res} shows that the BART baseline outperforms previous state-of-the-art models~\cite{Zhao2020Low-Resource,prabhumoye-etal-2019-towards} on all three tasks.
It demonstrates that both our improvements DoHA and CoDR perform better than our BART baseline on all metrics and for all three tasks.
Notably, we see an improvement of $19.7$ BLEU-4 points on the CMU\_DoG dataset compared to~\citet{Zhao2020Low-Resource} which was pre-trained on dialogue specific data; and an improvement on $8.9$ BLEU-4 points on the Wikipedia Update Generation compared to~\cite{prabhumoye-etal-2019-towards}.\footnote{We use NLG eval script for 
~\cite{prabhumoye-etal-2019-towards}}
We also see substantial improvements ($23.6$\% increase in BLEU-4 for CMU\_DoG) compared to the simple BART baseline for the three tasks.
In general, DoHA performs slightly better than CoDR on the three tasks.

\begin{table*}[h!]
\centering
\small{
\begin{tabular}{@{}l r r r @{\hskip 0.7in} r r r @{\hskip 0.7in} r r r@{}}
\multicolumn{1}{c}{\textbf{Task}} & \multicolumn{3}{c}{\textbf{BART v CoDR}}{\hskip 0.7in} & \multicolumn{3}{c}{\textbf{BART v DoHA}}{\hskip 0.7in} & \multicolumn{3}{c}{\textbf{DoHA v CoDR}} \\
\toprule
 & BART & NoPref & CoDR & BART & NoPref & DoHA & DoHA & NoPref & CoDR \\
\cmidrule{2-10}
 & \multicolumn{9}{c}{\tb{Wikipedia Update Generation}} \\
\cmidrule{2-10}
\ti{Closeness} & 33.3 & 36.7 & 30.0 & 25.5 & 46.7 & 27.8 & 32.2 & 42.2 & 25.6 \\
\ti{Relevance} & 18.9 & 54.4 & 26.7 & 24.4 & 45.6 & 30.0 & 33.3 & 38.9 & 27.8 \\
\toprule
& \multicolumn{9}{c}{\tb{CMU\_DoG}} \\
\cmidrule{2-10}
\ti{Closeness} & 15.6 & 58.8 & 25.6 & 30.0 & 42.2 & 27.8 & 33.3 & 44.5 & 22.2 \\
\ti{Relevance} & 22.2 & 43.4 & 34.4 & 23.3 & 42.3 & 34.4 & 34.4 & 42.3 & 23.3 \\
\toprule
& \multicolumn{9}{c}{\tb{Wizard of Wikipedia (seen)}} \\
\cmidrule{2-10}
\ti{Closeness} & 36.7 & 40.0 & 23.3 & 28.9 & 31.1 & 40.0 & 40.5 & 31.7 & 27.8 \\
\ti{Relevance} & 24.2 & 51.6 & 24.2 & 32.2 & 35.6 & 32.2 & 28.9 & 46.7 & 24.4 \\
\toprule
& \multicolumn{9}{c}{\tb{Wizard of Wikipedia (unseen)}} \\
\cmidrule{2-10}
\ti{Closeness} & 23.3 & 47.8 & 28.9 & 44.4 & 20.0 & 35.6 & 21.1 & 63.3 & 15.6 \\
\ti{Relevance} & 27.8 & 47.8 & 24.4 & 30.0 & 43.3 & 26.6 & 23.3 & 41.1 & 35.6 \\
\bottomrule
\end{tabular}
}
\vspace{-0.5em}
\caption{Human evaluation results depicting percentage of times a model was picked (NoPref=No Preference)}
\vspace{-1.0em}
\label{tab:gg_human_eval}
\end{table*}

\subsection{Human Evaluation}
\label{sec:gg_human_eval}
We follow the human evaluation guidelines mentioned in~\cite{prabhumoye-etal-2019-towards} and evaluate the system generated sentences on three dimensions: (1) closeness of the generated sentences to the references, (2) relevance of the generated sentences to the context and document, and (3) fluency of the generated sentences.

\paragraph{Closeness:} 
The automatic metrics like BLEU, METEOR, and Rouge-L may not be tolerant towards linguistic variations in generated outputs.
Hence, we perform a human evaluation to measures how accurately the generated sentence reflects the information in the reference.
The annotators are provided with the reference sentence and the generated outputs of two systems labeled $A$ and $B$ in a randomized order.
The annotators were instructed to ``\ti{Pick the option which is closest in meaning with the reference option.}''
The annotators could select system $A$ or $B$, or indicate that neither was preferred by picking the third option $C$.
This is a simple evaluation task though potentially biased toward the sole reference.

\paragraph{Relevance:} 
The reference sentence may not be the only correct sentence that fits the context.
This is especially true in dialogue generation tasks where contexts like \ti{``How are you?''} and \ti{``What was your favourite part of the movie?''} can have many correct responses that can be produced by grounding on the same document.
Hence, we measures whether the generated output contained salient information from the document written in a manner appropriate to the context.
The annotators are provided with the document $\B{d}_i$, the context $\B{c}_i$, and the outputs of the two systems $A$ and $B$, again in a random order.
They were instructed to \ti{``Pick the option which contains information from the document and fits the dialogue context coherently}''.
Note that the annotators don't have access to the reference in this evaluation.
Each judge had to consider whether the information fits with the context and also whether system-generated content could be supported by the document.

\begin{table*}[t]
\centering
\small{
\begin{tabular}{@{}p{1.1in}@{\hskip 0.1in} p{0.2in}@{\hskip 0.1in} p{1.5in}@{\hskip 0.2in} p{1.2in}@{\hskip 0.2in} p{1.6in}@{}}
\textbf{Error Class} & \textbf{\%} & \textbf{Chat context} & \textbf{Reference} & \textbf{Generation}\\
\toprule
\emph{Reference and generation are grounded} & 35 & 
the story is sounding even more  interesting.  who do you think had the most growth or development in the film? & 
but there's a twist in that \textcolor{darkgreen}{anna has been deceived by the man she was in love with, he was really plotting to take over the kingdom from the sisters the whole time.} & 
but there's a twist in that\textcolor{darkgreen}{anna is actually in love with a man who is actually a bad guy. he was actually trying to take over the kingdom from the sisters after elsa injured anna} and then their parents died.
\\
\midrule
\emph{Reference is grounded but generation is not} & 3 & 
hello! & 
have you heard about \textcolor{darkgreen}{the post directed by spielberg?} & hello! \\
\midrule
\emph{Reference is not grounded but generation is} & 5 & 
yes i agree and how meryl can depict the first female publisher. that would be interesting to see the dynamic of how it effected her career as well & 
yes, for a woman to have a successful career in those times was not an easy task, for sure. & 
definitely. i wonder how \textcolor{darkgreen}{mcnamara} felt when he saw the documents and then \textcolor{darkgreen}{told ellsberg that the war was hopeless.} \\
\midrule
\emph{Reference and generation are not grounded} & 43 & 
the shark was so realistic & 
yea i live near a beach and it made me not ever want to go swimming & 
i never thought of it that way, and i agree, it made me terrified to go into the ocean for awhile.\\
\bottomrule
\end{tabular}
}
\vspace{-0.5em}
\caption{Error analysis for the CMU\_DoG dataset (only shown for outputs with low Rouge-L score). Text in \textcolor{darkgreen}{green} indicates grounded information.}
\vspace{-1.0em}
\label{tab:gg_error_analysis_cmu}
\end{table*}

\paragraph{Fluency:} 
Finally, we evaluate the fluency of the generated sentences on a scale of 1 (unreadable) to 4 (perfect) as is described in~\cite{zhou-etal-2018-dataset}.

Human evaluation was conducted on Amazon Mechanical Turk.
We conduct $3$ comparative studies between the BART, CoDR and DoHA outputs.
Each worker was asked to annotated $10$ pairs of sentences.
We added one control pair among them i.e for $1/10$ pairs, both the sentences were exactly the same.
If a worker provides wrong judgement for the control pair then their annotations were discarded.
For each dataset we have total $540$ comparative judgements and $90$ sentences of each of the models marked for fluency.

Table~\ref{tab:gg_human_eval} shows the results of the human evaluation on closeness and relevance. 
The closeness results show that all the three models BART, CoDR and DoHA generate sentences that are close to the reference, although CoDR and DoHA outperform BART in most cases.
Interestingly, the relevance results for Wikipedia Update Generation and CMU\_DoG datasets show that CoDR and DoHA generate content that is grounded in the document as opposed to BART. 
BART baseline generates sentences that are fluent and close to the reference but does not ground in the content of the document as compared to CoDR and DoHA.
The `No Preference' is generally opted over any of the models which is further discussed in \Sref{sec:gg_discuss}.
For the relevance comparison, annotators have to read a large document to figure out if the generated information is present in the document or not. 
This can make the annotations noisy especially for Wizard of Wikipedia dataset which has $7$ passages as grounding document.

Since both CoDR and DoHA are also BART-based models, the fluency for all three of them is very high and close to each other (BART=$3.64$, CoDR=$3.71$, DoHA=$3.66$).

\paragraph{CoDR and DoHA:}
The DoHA model still uses the content driven representations ($\B{h}_d$ and $\B{h}_c$).
The main difference is that in CoDR model we concatenate $\B{h}_d$ and $\B{h}_c$ and pass it to the decoder but for DoHA we pass  $\B{h}_d$ and $\B{h}_c$ separately to the decoder. 
DoHA has an additional attention layer to focus on the representation of the document $\B{h}_d$ only. 
In this loose sense, DoHA is CoDR plus additional parameters in attention layer to focus on $\B{h}_d$. 
DoHA performs marginally better than CoDR in automated metrics. 
But qualitatively (human evaluation) DoHA produces higher quality outputs as compared to CoDR. 
Table~\ref{tab:gg_human_eval} shows DoHA performing better than CoDR on all but one case. 

\section{Analysis and Discussion}
\label{sec:gg_discuss}

We manually inspect the outputs of the CoDR model on the development set of CMU\_DoG and Wikipedia Update Generation dataset to understand the their quality.
We inspect 60 samples in each dataset which have Rouge-L score $< 60$.
These are chosen such that we have 10 samples in each of the 6 buckets of Rouge-L score (buckets are range of 10 points: 0-9, 10-19, 20-29, 30-39, 40-49 and 50-59).
We analyse the generated outputs along the two aspects of appropriateness of the generation to the context and its grounding in the document.

\paragraph{CMU\_DoG:} We find that $52/60$ ($86.7$\%) responses were appropriate to the given chat context.
These $52$ responses are further categorized in Table~\ref{tab:gg_error_analysis_cmu}.
We found that for about $90$\% of samples, if the reference is grounded then the generation is also grounded and if the reference is not grounded then the generation is not grounded.
Further inspection shows that references are not grounded if they are follow up questions, opinions or experiences that are shared in the conversation.
In most of these cases, the context dictates if the response should be grounded or not grounded in the document.
Since, all of the generated responses in this category are appropriate to the context suggests that these conversational subtleties are not captured by automated evaluation metrics and are given a low score.
We also observe a few data artifacts like the mapping of the Wikipedia sections and the chat context is noisy for this dataset.
This can be easily resolved by providing all the previous passages of the conversation as grounding to the model.
We would also like to note that this dataset was collected under two scenarios: (1) both the people in the conversation have access to the document, and (2) only one person has access to the document.
But this distinction is not made in modeling the task.
The noise in the dataset can be reduced by modeling only the users that have access to the document in the conversation (similar to Wizard of Wikipedia where only the wizard is modeled).

\begin{table*}[t]
\centering
\small{
\begin{tabular}{@{}p{1.7in}@{\hskip 0.1in} p{0.2in}@{\hskip 0.1in} p{1.8in}@{\hskip 0.2in} p{1.5in} p{0.2in}@{}}
\textbf{Error Class} & \textbf{\%} & \textbf{Reference} & \textbf{Generation} & \tb{R}\\
\toprule
\emph{Linguistic Variation:} Reference and generation are grounded and generation is appropriate but a linguistic variation of the reference or an alternate appropriate update. & 43 & December 12 - The Smiths play Brixton Academy, their last ever gig before their dissolution. & December 12 - The Smiths perform their final show, at Brixton Academy in London. & 41 \\
\midrule
\emph{Partial Hallucination:} Reference and generation are grounded but generation is either missing or hallucinates some information & 23 & America Online and Prodigy (online service) offered access to the World Wide Web system for the first time this year, \textcolor{red}{releasing browsers that made it easily accessible to the general public.} &  The World Wide Web was first introduced on January 17, \textcolor{red}{1995} on Prodigy. & 17 \\
\midrule
\emph{Incoherent Reference:} The reference does not coherently follow the context & 22 & ``The Naked Ape'', by Desmond Morris, is published. &  Zoologist Desmond Morris publishes ``The Naked Ape''. & 26 \\
\midrule
\emph{Incorrect:} The generation is either not appropriate or is not grounded (completely hallucinates the information). & 7 & The year 2000 is sometimes abbreviated as ``Y2K'' (the ``Y'' stands for ``year'', and the ``K'' stands for ``kilo-'' which means ``thousand''). & \textcolor{red}{The Y2K conspiracy theory claimed that a secret nuclear attack by the United States on 2 January 2000 was planned to begin World War 2.} & 9 \\
\midrule
\emph{Reference is not grounded} & 5 & \textcolor{red}{This was achieved under dead calm conditions as an additional safety measure, whereas the Wrights flew in a 25 mph+ wind to achieve enough airspeed on their early attempts.} & \textcolor{red}{This was verified by a video crew present at the test flight.} & 14 \\
\bottomrule
\end{tabular}
}
\vspace{-0.5em}
\caption{Error Analysis for Wikipedia Update Generation task (R denotes Rouge-L score. Text in \textcolor{red}{red} indicates hallucinated or missing information.)}
\vspace{-1.0em}
\label{tab:gg_error_analysis_wiki}
\end{table*}

\paragraph{Wikipedia Update Generation:} The error analysis for this task is shown in Table~\ref{tab:gg_error_analysis_wiki}.
For $5$\% cases, the reference itself is not grounded in the document.
The remaining $95$\% cases are further classified into $4$ error categories.
About $85$\% times, the generation is either completely or partially grounded if the reference is grounded.
$43$\% generations are grounded in document but are linguistic variations of the reference or could be alternate updates to the context.
Yet, these are scored low on the Rouge-L metric revealing the inadequacy of the automated metrics.
For $23$\% cases the generation partially hallucinates some information or misses some information present in the reference.
$22$\% times the reference itself does not seem to coherently fit the context.
This is primarily observed for Wikipedia pages that are in the form of a list like \ti{1340s} and \ti{Timeline of DC Comics (1950s)}.
Yet, for $50$\% of the \ti{Incoherent Reference} cases, the generation is grounded in the document and very close to the reference (like the example in Table~\ref{tab:gg_error_analysis_wiki}).
Only for $7$\% of the cases, the generation is completely incorrect and hallucinates all of the information.
Future work can focus on improving the error in the \ti{Incorrect} and \ti{Partial Hallucination} error classes.

\vspace{-0.25em}
\paragraph{Reference Comparison:} With the insights from manual inspection, we performed another comparative study with human judges (on Amazon Mechanical Turk).
This was to understand how our models perform in comparison with the reference.
The judges are instructed to ``\ti{Pick the option that is most appropriate to the given context}''.
We annotated $100$ samples for each DoHA and CoDR model in comparison with the reference on the CMU\_DoG and Wikipedia Update Generation datasets.
We perform two separate comparative experiments: Reference vs CoDR and Reference vs DoHA.  
The results in Table~\ref{tab:gg_ref_compare} show consolidated results for the two models.
It shows the total number of times reference was selected, the total number of times ‘No Pref’ was selected or the total number of CoDR or DoHA was selected.
It demonstrates that our models produce appropriate outputs which can be used as alternate responses/updates.
Our models are preferred over the reference in both the tasks suggesting that the automated evaluation is insufficient and the sole reference should not be considered as the only correct response to the context.

\section{Related Work}
\label{sec:gg_rel_work}

Generation grounded in document has been studied through a large body of summarization work~\cite{rush-etal-2015-neural,nallapati2016abstractive} and similar tasks such as headline generation~\cite{tan2017neural}.
Multiple new works have extended this research in new directions; Wikipedia Update Generation~\cite{prabhumoye-etal-2019-towards} introduces the task of generating an \emph{update} to the Wikipedia context based on a news document; Wikipedia article generation~\cite{WikiGen2018} introduces the task of generating an entire Wikipedia article based on multiple documents; Text Editing by Command~\cite{faltings2020text} introduces the task of generating a particular type of Wikipedia edit conditioned on a command provided in natural language and a grounding consisting of snippets of 200 web page results. 

Parallely, new tasks have also emerged focusing on document grounding for dialogue response generation~\cite{zhou-etal-2018-dataset,dinan2018wizard}.
\citet{Zhao2020Low-Resource} explore this task in low-resource setting and use pre-training along with a disentangled decoder.
The disentangled decoder consists of a context processor, knowledge processor and a language model.
A dialogue manager is used to combine the vocabulary distributions provided by these three components.
\citet{zhao2020knowledgegrounded} propose a knowledge selection module integrated with pre-trained language models for this task.

\citet{cao-etal-2020-pretrained} use pre-trained language model GPT-2~\cite{radford2019language} and explore various attention fusion techniques for persona-based dialogue generation~\cite{zhang-etal-2018-personalizing,dinan2020second}.
Our DoHA technique also introduces an additional attention multi-head but does not use any additional weights to fuse attention heads.
Similarly,~\citet{junczys-dowmunt-grundkiewicz-2018-ms} use an additional attention multi-head in transformer architecture for automatic post-editing task.
We demonstrate how attention can be enhanced in pre-trained models.
The CoDR model fuses the representations of the document and the context in the decoder which is inspired by the fusion-in-decoder model in open-domain QA~\citep{fusion-in-decoder}.
Although~\citet{Bruyn2020BARTFK} introduce the usage of BART for knowledge grounded dialogues, it is primarily from the perspective of improving knowledge retrieval. 
We provide benchmark BART numbers (Table~\ref{tab:gg_main_res}) for the generation task.
~\citet{prabhumoye-etal-2020-exploring} provide a schema containing five modules which can be changed to control the generation process.
While~\citet{Zhao2020Low-Resource} modify the external input and the output module, we focus on the external input and the generator module of the pre-trained language model.

\begin{table}[t]
\centering
\small{
\begin{tabular}{@{}l r r r@{}}
\textbf{Dataset} & \textbf{Ref} & \textbf{NoPref} & \textbf{DoHA/CoDR} \\
\toprule
Wikipedia & 33.9 & 28.3 & 37.8 \\
CMU\_DoG & 22.8 & 45.6 & 31.6 \\
\bottomrule
\end{tabular}
}
\vspace{-0.5em}
\caption{
Comparison with reference (Ref) in \%age}
\vspace{-1.0em}
\label{tab:gg_ref_compare}
\end{table}



\section{Conclusion and Future Work}
\label{sec:gg_conc}

This paper proposes two novel improvements for document grounded generation and provides a stronger baseline.
This paper demonstrates how similar modeling techniques could be used for two previously separately modeled tasks.
Our proposed models outperform the previous techniques and the new stronger baseline on automated metrics and human evaluation for the three datasets discussed in the paper.
We present a comprehensive manual inspection which reveal certain data artifacts and provides us with insight on how to model these tasks in future.
Particularly, future work can focus on designing better evaluation metrics which don't penalize linguistic variations in generation.
Better models can also be constructed to focus on cases of partial hallucination or incorrect responses.

\section{Ethical Considerations}
\label{sec:gg_ethics}

The intended use of the models proposed is to aid the NLG systems in generating content-rich text.
Note that this does not imply that the models generate factually correct text.
The generation entirely depends on the information in the document provided.
If the document itself is factually incorrect then the generation would be grounded in false content and hence generate inaccurate text.

We hope that this technology is used for socially positive applications like building trust of users in dialogue systems like Alexa, Siri and Google Home by providing users with credible information.
This work has specifically focused on two datasets of dialogue response generation with the aim that this research not only helps in generating responses which contain useful information but also increase credibility of responses by disclosing the source of information.
If dialogue systems base their responses on certain sources of information then they can potentially disclose the source of information to the user.
The user then has the agency to make informed decision about trusting the system responses or not.

Additional generations are shown in Appendix (\Sref{sec:appendixB}).
Table~\ref{tab:gg_wrong_gen_wiki} and~\ref{tab:gg_wrong_gen_wizard} in  Appendix \Sref{sec:appendixB} show the potential misuses of models trained on this task.
For both the experiments, a few news articles were hand selected and relevant context was selected from a chosen Wikipedia article. 
In case of Table~\ref{tab:gg_wrong_gen_wizard}, the context was curated by hand.
Interestingly, the tables also shows the sensitivity of the trained model to the document information.
It consists of the same context but different documents were provided as inputs to the model.
The generated outputs are different for each document.

\section*{Acknowledgements}
This work was supported in part by ONR Grant N000141812861 and NSF IIS1763562.
We are grateful to Semih Yavuz and Caiming Xiong for valuable discussions at earlier stages of this work.
We would like to thank Srinath Reddy Meadusani for his technical support throughout the project.

\bibliography{anthology,custom}
\bibliographystyle{acl_natbib}

\clearpage
\appendix

\section{Appendix A}
\label{sec:appendixA}
\subsection{Implementation Details}
We use the transformer toolkit~\cite{Wolf2019HuggingFacesTS} to implement the baseline and both CoDR and DoHA models.\footnote{The results are subject to changes in the codebase of the toolkit. Note that we will release our code and trained models to ensure reproducbility of results.}
Both DoHA (\Sref{sec:gg_doha}) and CoDR (\Sref{sec:gg_codr}) have the same dimensions and architecture of the BART model~\cite{lewis-etal-2020-bart}.
For the DoHA model, we initialize \ti{CrossAttention\_Doc} with same pre-trained weights of \ti{CrossAttention}.
Hence, the layer size of the \ti{CrossAttention\_Doc} multi-head is the same as the layer size of \ti{CrossAttention} multi-head in BART.
Table~\ref{tab:dd_data_lens} shows the maximum sequence lengths used for all the three datasets for both source and target.
The data statistics are shown in Table~\ref{tab:gg_data_stats}.\footnote{We try to closely follow the processing of the original papers for each of the three datasets.}
We experimented with two learning rates $5$e-$5$ and $2$e-$5$. 
We report numbers for the best trained models in each case.
Specifically, we report numbers with $5$e-$5$ learning rate for DoHA and CoDR models on the CMU\_DoG dataset and the BART baseline for all the three datasets.
For Wikipedia Update Generation and Wizard of Wikipedia dataset, we choose the DoHA and CoDR models trained with $2$e-$5$ learning rate.
We maintain a common environment (in terms of GPU, operating system, Pytorch version and transformer version) to run all the experiments.
We train all the models for $25$ epochs.

\citet{Zhao2020Low-Resource} numbers are directly taken from the paper as the pre-trained model or the generated outputs are not available. We use the same data splits and evaluation toolkits for comparable setting. Hence, Rouge-L and Meteor values are not available for this model.
The BLEU, Meteor and Rouge-L numbers are different from
~\cite{prabhumoye-etal-2019-towards} due to the usage of different tool-kits in measuring their values.
 
\begin{table}[h!]
\centering
\small{
\begin{tabular}{@{}l r r@{}}
\textbf{Dataset} & \textbf{Source Len} & \textbf{Target Len} \\
\toprule
Wikipedia Update Generation & 1024 & 128 \\
CMU\_DoG dataset & 512 & 128 \\
Wizard of Wikipedia & 900 & 40 \\
\bottomrule
\end{tabular}
}
\caption{Sequence Lengths}
\label{tab:dd_data_lens}
\end{table}

\begin{table}[h!]
\centering
\small{
\begin{tabular}{@{}l r r r@{}}
\textbf{Dataset} & \textbf{Train} & \textbf{Dev} & \textbf{Test} \\
\toprule
Wikipedia Update Generation & 580.0k & 6.0k & 50.0k\\
CMU\_DoG & 72.9k & 4.8k & 13.2k \\
Wizard of Wikipedia & 166.7k & 17.7k & 8.7k\\
\bottomrule
\end{tabular}
}
\caption{Dataset Statistics}
\label{tab:gg_data_stats}
\end{table}

\begin{figure}[t]
\centering
\includegraphics[width=\linewidth]{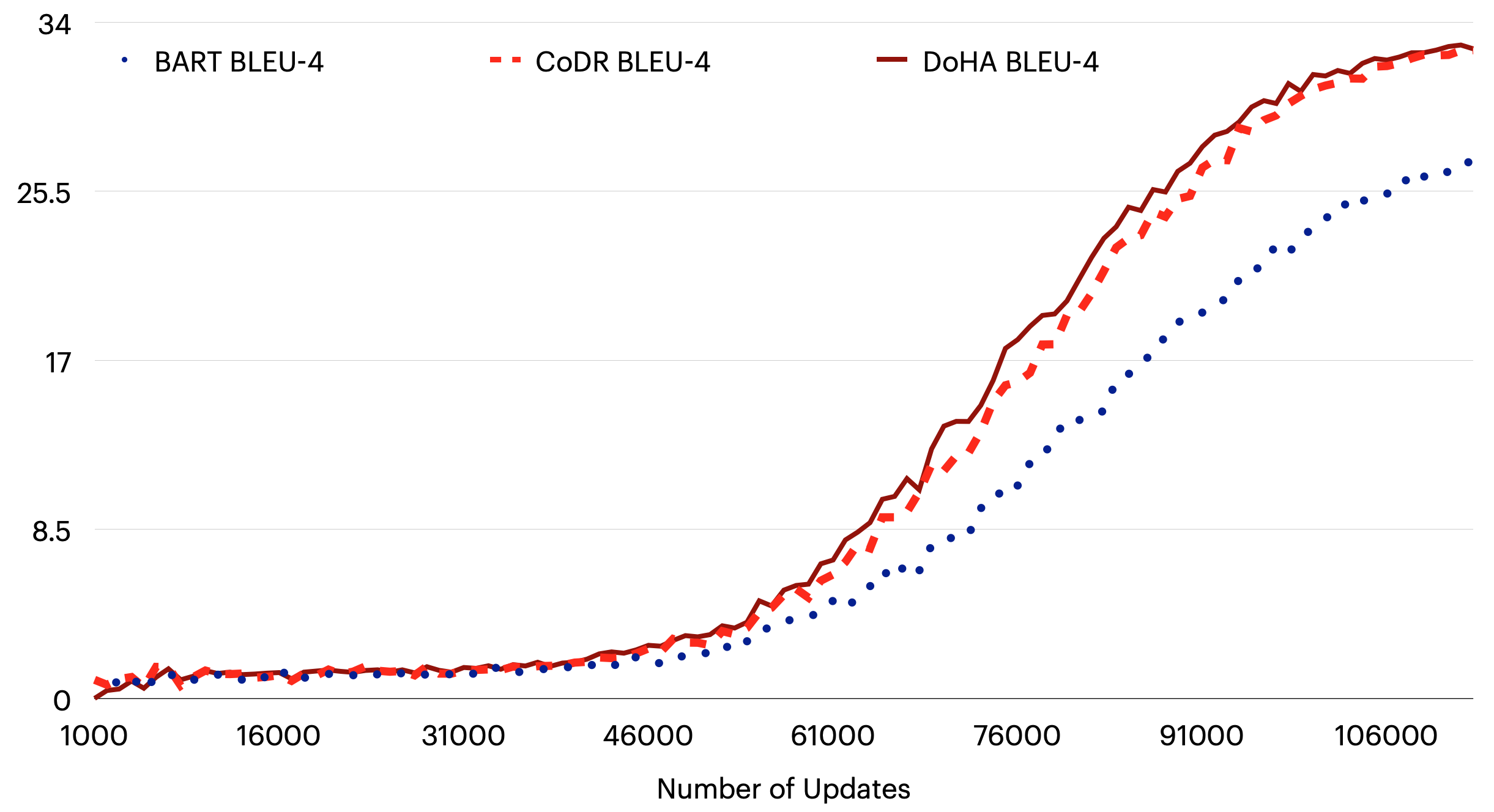}
\caption{Convergence of CMU\_DoG development data on the automated metric.}
\label{fig:gg_convergence_cmu_dev}
\vspace{-1em}
\end{figure}

\begin{figure}[h!]
\centering
\includegraphics[width=\linewidth]{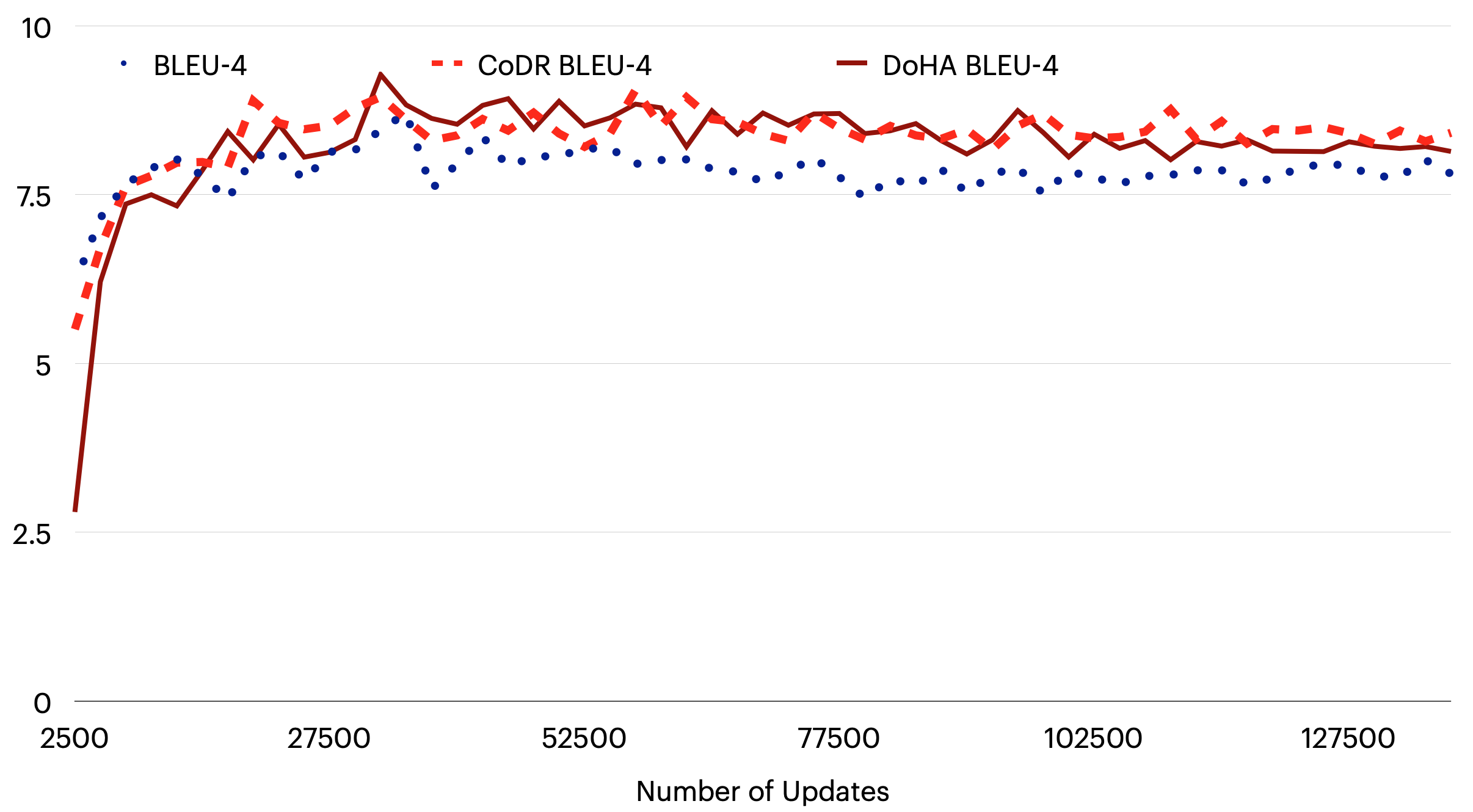}
\caption{Convergence of CMU\_DoG development data on the automated metric.}
\label{fig:gg_convergence_wiz_dev}
\vspace{-1em}
\end{figure}

\paragraph{Convergence:}
Figures~\ref{fig:gg_convergence_cmu_dev} and~\ref{fig:gg_convergence_wiz_dev} shows the convergence of the baseline BART model in comparison with the CoDR and DoHA models on the development sets of CMU\_DoG and Wizard of Wikipedia respectively.
We observe that at same number of updates, DoHA and CoDR perform better than BART.
This is especially relevant for big datasets like Wikipedia Update Generation which take 15 days to complete $25$ epochs.

\section{Appendix B}
\label{sec:appendixB}

We hand selected a few news articles which contained inaccurate information.
We provided relevant Wikipedia context and generated an update for the context using the trained model on Wikipedia Update generation dataset. 
The generations are shown in Table~\ref{tab:gg_wrong_gen_wiki}.
Additionally, we provided relevant dialogue context to the model trained on Wizard of Wikipedia dataset (CMU\_DoG was not chosen because it is trained only on movie conversations) and the responses generated are shown in Table~\ref{tab:gg_wrong_gen_wizard}.
Note that this shows the potential misuse of the task design.

\begin{table*}[t]
\centering
\small{
\begin{tabular}{@{}p{1.3in}@{\hskip 0.2in}  p{2.5in}@{\hskip 0.2in} p{1.8in}}
\tb{Document} & \tb{Context} & \tb{Generation} \\
\toprule
\url{https://www.bradenton.com/latest-news/article237480339.html}
& 
Climate change includes both the global warming driven by human emissions of greenhouse gases, and the resulting large-scale shifts in weather patterns. Though there have been previous periods of climatic change, since the mid-20th century the rate of human impact on Earth's climate system and the global scale of that impact have been unprecedented.
& 
Though most scientists agree that a ``global'' climate change is occurring, climate change remains a contentious topic and many scientists, climate scientists, and even climate skeptics believe that it is a hoax. \\
\midrule
\url{https://tinyurl.com/3r5runb}
& 
Climate change includes both the global warming driven by human emissions of greenhouse gases, and the resulting large-scale shifts in weather patterns. Though there have been previous periods of climatic change, since the mid-20th century the rate of human impact on Earth's climate system and the global scale of that impact have been unprecedented. 
&
Some scientists dismiss the existence of climate change. \\
\midrule
\url{https://tinyurl.com/y634czdp}
& Climate change includes both the global warming driven by human emissions of greenhouse gases, and the resulting large-scale shifts in weather patterns. Though there have been previous periods of climatic change, since the mid-20th century the rate of human impact on Earth's climate system and the global scale of that impact have been unprecedented.
& The Scientific opinion on climate change is that planetwide warming has occurred at least three million years ago.
\\
\bottomrule
\end{tabular}
}
\caption{Outputs generated by DoHA model for a few hand selected news articles. The DoHA model is trained on Wikipedia Update Generation dataset.}
\label{tab:gg_wrong_gen_wiki}
\end{table*}

\begin{table*}[t]
\centering
\small{
\begin{tabular}{@{}p{1.0in}@{\hskip 0.2in}  p{2.0in}@{\hskip 0.2in} p{2.5in}}
\tb{Document} & \tb{Context} & \tb{Generation} \\
\toprule
 \url{https://tinyurl.com/yxedlapz}
 &
What do you think of the rising sea levels?
&
I think it's crazy that people believe that the oceans will rise and yet people are investing \$15 million to live on islands...\\
\midrule
\url{https://tinyurl.com/y3u6fhpo}
& 
Do you think the Affordable Care Act is helping the general public?
& 
It has some great benefits.It has actually resulted in millions of Americans losing their health insurance plan. That is not something former VP Biden claimed. \\
\midrule
\url{https://tinyurl.com/3r5runb}
& 
What do you think of the rising sea levels?
&
I think the rise of sea levels is a big money making hoax. Why do you think the world is heating up? Any experts? \\
\bottomrule
\end{tabular}
}
\caption{Outputs generated by DoHA model for a few hand selected news articles. The DoHA model is trained on Wizard of Wikipedia dataset.}
\label{tab:gg_wrong_gen_wizard}
\end{table*}

\end{document}